\documentclass[10pt,twocolumn,letterpaper]{article}

\usepackage{iccv}
\usepackage{times}
\usepackage{animate}
\usepackage{epsfig}
\usepackage{graphicx}
\usepackage{amsmath}
\usepackage{amssymb}
\usepackage{mathtools}
\usepackage[mathscr]{euscript}
\usepackage{xcolor}
\usepackage{subcaption}
\usepackage{pifont}

\usepackage[pagebackref=true,breaklinks=true,letterpaper=true,colorlinks,bookmarks=false]{hyperref}
\hypersetup{
    colorlinks=true,
    linkcolor=teal,
    citecolor=green,
    filecolor=magenta,      
    urlcolor=violet,
}

\iccvfinalcopy 

\def\iccvPaperID{15} 

\ificcvfinal\fi
\begin{document}

\title{Class Feature Pyramids for Video Explanation}

\author{Alexandros Stergiou$^{1}$ \qquad Georgios Kapidis$^{1}$ \qquad Grigorios Kalliatakis$^{2}$ \qquad Christos Chrysoulas$^{3}$ \\Ronald Poppe$^{1}$ \qquad Remco Veltkamp$^{1}$\\
\hfill\linebreak[3]\\
\begin{tabular}{c c c}
$^{1}$Utrecht University & $^{2}$University of Essex & $^{3}$London South Bank University \tabularnewline
Utrecht, The Netherlands & Colchester, United Kingdom & London, United Kingdom
\end{tabular}
}

\maketitle

\begin{abstract}
   Deep convolutional networks are widely used in video action recognition. 3D convolutions are one prominent approach to deal with the additional time dimension. While 3D convolutions typically lead to higher accuracies, the inner workings of the trained models are more difficult to interpret. We focus on creating human-understandable visual explanations that represent the hierarchical parts of spatio-temporal networks. We introduce \textit{Class Feature Pyramids}, a method that traverses the entire network structure and incrementally discovers kernels at different network depths that are informative for a specific class. Our method does not depend on the network's architecture or the type of 3D convolutions, supporting grouped and depth-wise convolutions, convolutions in fibers, and convolutions in branches. We demonstrate the method on six state-of-the-art 3D convolution neural networks (CNNs) on three action recognition (Kinetics-400, UCF-101, and HMDB-51) and two egocentric action recognition datasets (EPIC-Kitchens and EGTEA Gaze+).\footnotemark
   \footnotetext{Our code is available at \hyperlink{https://git.io/fjDCW}{https://git.io/fjDCW}}
\end{abstract}
\begin{figure}[htb]
    \centering
    \animategraphics[loop,autoplay,width=.7\columnwidth]{8}{CFVP_frames/frame-}{01}{49}
\caption{\textbf{Class Feature Pyramid.} A top-down view of class-specific feature information at each layer, represented as a hierarchical pyramid. The figure is created for an \textit{abseiling} video  from Kinetics-400 on a 3D-ResNet152. Figure is animated.\vspace{-1mm}}
\label{fig:fig1}
\end{figure}

\section{Introduction}
Deep neural networks have revolutionized many domains, \textit{e.g.}, image recognition, speech recognition and knowledge discovery \cite{krizhevsky2012imagenet,lecun2015deep,lecun2012efficient}. With the success of those modern neural networks, comes the need to explain their decisions -- including understanding how they will behave in the real world, detecting model bias, and for scientific curiosity. However, their nested non-linear structure makes them highly non-transparent, \textit{i.e.}, neurons and layers in the network structure that significantly contribute to a class and instance-specific features cannot be accurately determined solely by the model's decisions. Therefore these models are typically regarded as \textit{black boxes}.

We focus on convolutional neural networks (CNNs), deep neural networks that use kernels to process the input data. Initial efforts to provide visual explanation of 2D CNNs followed the approach of learning a separate linear layer with global feature representations \cite{zeiler2014visualizing}. This approach has been widely used in order to discover the image regions that deep and complex convolutional architectures find as most informative \cite{montavon2018methods,zhou2016learning}. The functionality of such methods stretches beyond simple information representation, providing additional benefits for various tasks. Recent works have studied class activations for further empowering models and as the basis for teacher-student networks \cite{jetley2018learn,lu2017knowing,woo2018cbam,zagoruyko2016paying}. Others have focused on utilizing feature information for 
restricting model capacity through ranking of how globally informative features are \cite{cheng2018recent,luo2017thinet}.

Although impressive progress has been made in the domain of visual explanations in the 2D image domain \cite{anne2018grounding,hendricks2016generating,samek2017explainable,shrikumar2016not}, visualization techniques for 3D convolution operations are scarce. This is mainly due to the additional time dimension of spatio-temporal frame sequences.


In this work, we bridge the gap between video classification and the interpretability of 3D-CNNs with a novel method \textit{Class Feature Pyramids}: a plug-and-play visualization method specifically designed for 3D-CNNs. Class Feature Pyramids enable \textit{back-stepping}\footnotemark ~across multiple network layers and generating a class dependency graph that provides a tree-like hierarchy of the most informative features and their neural connections to feature activations in the preceding and succeeding layers. This enables the creation of a pyramid-like representation of class-informative layer features as seen in Figure~\ref{fig:fig1}. 

\footnotetext{We define the exploration of cross-layer features as \textit{back-stepping} through the network in order to construct an association between high-level and low-level features.}

Our contributions are summarized as follows:
\begin{itemize}
\item We introduce Class Feature Pyramids to improve the interpretability of 3D-CNNs by finding class-specific features across all network's layers and neglecting 3D convolutional kernel's spatio-temporal locality. Class Feature Pyramids are developed as a generic, plug-and-play visualization technique that can be used for any architecture that includes different types of convolutions or connections without any adjustments in the network architecture.
\item We extensively study the properties of Class Feature Pyramids on visualizing layer-wise features as a concatenation of multiple extracted features and individual activation maps, based on their connection with other features with high activations from a previous layer.
\item We apply Class Feature Pyramids to six top-performing spatio-temporal models in five benchmark action recognition datasets Kinetics-400 \cite{carreira2017quo}, UCF-101 \cite{soomro2012ucf101}, and HMDB-51 \cite{kuehne2011hmdb} and egocentric action recognition datasets EPIC-kitchens \cite{damen2018scaling} and EGTEA Gaze+ \cite{li_eye_2018}. We show that for action recognition, our method helps to uncover the time-space regions that networks focus on at different depths, while for egocentric action recognition, our method exposes the salient areas of the hand and object movements that the network associates with different action classes.
\end{itemize}

The remainder of the paper is structured as follows. We proceed with a discussion of related work on visualization techniques for CNNs. We introduce Class Feature Pyramids in Section~\ref{sec:sec3}. A quantitative and qualitative analysis of their performance appears in Sections~\ref{sec:section4} and \ref{sec:section5}, respectively. We conclude in Section~\ref{sec:section6}.

\section{Related work}

While modern CNNs involve high complexity\cite{berman2019multigrain,huang2018gpipe,tan2019efficientnet,Zoph_2018_CVPR}, the connection between model architectural evaluation and interpretability has become  closely associated with the overall model comprehensibility \cite{guidotti2018survey}. In most works that focus on network explainability, general feature information is extracted and presented by methods that only target the most prevalent high-level features learned by a network \cite{selvaraju2017grad,simonyan2013deep,van2014accelerating,zeiler2014visualizing}. They do not provide an extensive approach that can accommodate features used in different parts of the architecture. The rise of model complexity further indicates that explanation methods that solely consider the output of the entire model do not provide sufficient information when comparing different large architectures.

One of the first attempts to directly evaluate the importance and necessity of different parts of the network was that of Springenberg \etal \cite{springenbergstriving}. Their \textit{guided back-propagation} method made use of deconvolutions without the need of sub-sampling information in order to simulate a backward pass through the network. This also excluded negative gradients during the backward pass which relates to information from neural weights that decrease the high-layer activations. Through this, models can be decomposed to provide an understanding of the pixel-wise information extracted by multi-layer architectures \cite{bach2015pixel}. The proposed \textit{Layer-wise Relevant Propagation} (LRP) \cite{lapuschkin2016lrp} further enhances the cross-layer activation search by relating image parts to the corresponding class. Supplementary to LRP, \textit{DeepLIFT} \cite{shrikumar2017learning} used a score-keeping technique to compare neural activations through back-propagating over multiple network layers. The \textit{DeepLIFT} approach primarily focuses on understanding the importance of cross-layer connections through positive or negative score assignment. Other works on back-propagating information include optimization in order to find the activations that excite specific neurons through iteratively calculating the derivatives for permuting the neuron input \cite{olah2017feature}.

For video recognition, advances in explaining the features learned by spatio-temporal models have been scarce. Apart from the increased complexity in terms of the represented information in clips, visual descriptions in video models have been a challenging task also because of the lack of a fixed strategy based on which spatio-temporal models are built. Ways of representing the time dimension include Two-Stream networks \cite{simonyan2014two} that additionally use optical flow for capturing movement \cite{diba2017deep,park2016combining,wang2016temporal}, while others make use of recurrent cells \cite{gammulle2017two,liu2016spatio,zhu2016co}. One of the most dominant approaches has been the extension of 2D convolutions to 3D convolutions by including time as an additional dimension \cite{baccouche2011sequential,ji20133d}. 3D convolutions have yielded significant interest in the video recognition field \cite{carreira2017quo,ghadiyaram2019large,tran2018closer,tran2019video,feichtenhofer2018slowfast}. Since the popularization of 3D-CNNs for video recognition, some early attempts have been made for visualizing the activations of the learned features. An approach extending LRP to space-time volumes was introduced by Anders \etal \cite{anders2018understanding} in which a border effect was used for determining the objects and parts in frames that the network focuses on. 

In this paper, in order to enable a hierarchical representation of kernel layer activations in 3D-CNNs, we propose \textit{Class Feature Pyramids} for discovering kernel correspondence on the basis of how informative kernel activations are considered for classes. The visualization of kernel activations are based on the \textit{Saliency Tubes} approach \cite{stergiou2019saliency} which uses the spine interpolate of spatio-temporal kernel activations in order to create a representation conjoined with the used clip.

\section{Discovery of prominent spatio-temporal feature combinations}
\label{sec:sec3}

A number of previous works \cite{chattopadhay2018grad,stergiou2019saliency} have focused on creating representations of the regions in space and time that 3D-CNNs focus on when considering a particular class instance. These regions are class-specific and correspond to the activation maps produced in the last convolutional operation of the network. A shortfall of these methods is the fact that visually similar classes are salient in the same spatio-temporal region because the majority of their activations are the same, while only a small number of activations are distinct to a specific class. \textit{Class Feature Pyramids} explicitly target such features and their corresponding hierarchy, based on activations from previous layers of the network.

\begin{figure*}[!ht]
\centering
\includegraphics[width=\textwidth]{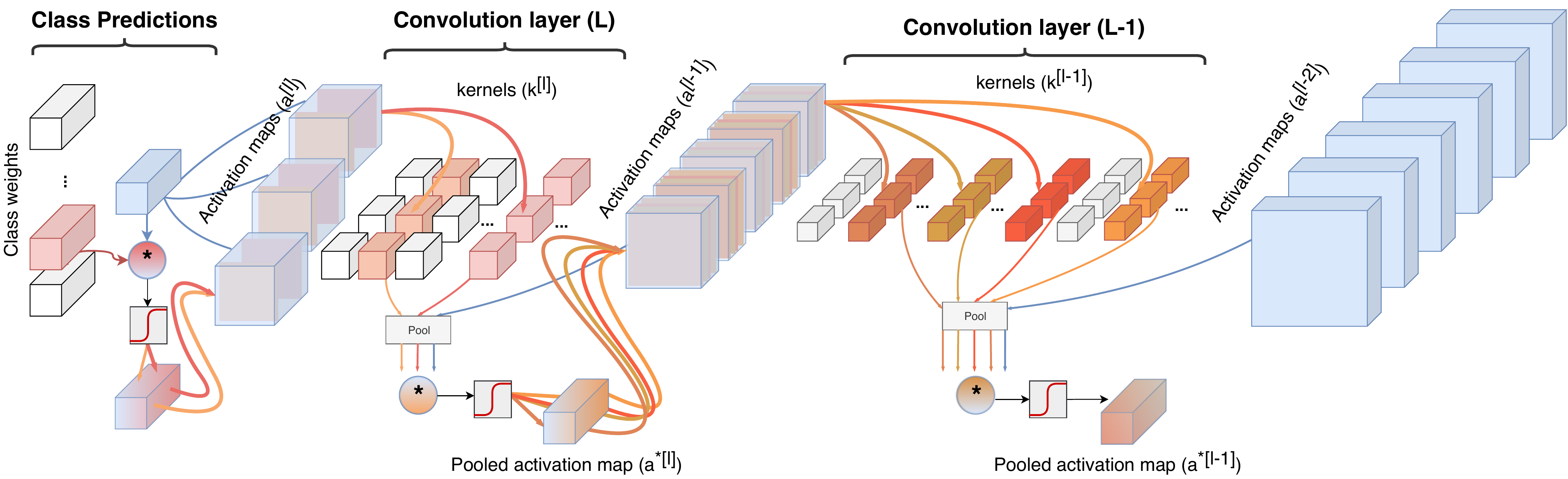}
\caption{\textbf{Back-step process for discovering significant class-specific features.} Layer weights and input activation maps are pooled to create global representations of the information in each space-time volume, alleviating the locality of convolutions. The selected kernels are multiplied element-wise to create new pooled class-based activation maps specific to only the selected features in the layer that exhibit large contribution for the specific class. The location of the highest activation in the class-based activations is found through a sigmoid applied over the volume. Features in a previous layer are iteratively discovered for each high activation in the current layer, thus creating a class feature hierarchy.}
\label{fig:fig2}
\end{figure*}

In this section, we will describe how cross-layer feature associations are found and how different networks can be back-propagated in a hierarchical way given their architectural building blocks. In Section~\ref{sec:section3.1} and through Figure~\ref{fig:fig2} we demonstrate how class information can be back-propagated through earlier layers. In Sections~\ref{sec:section3.2} and~\ref{sec:section3.3}, we provide an overview of how cross-layer feature dependencies are discovered. A description of how Class Feature Pyramids are used over different networks is found at Section~\ref{sec:section3.4}.

\subsection{Information propagation from class prediction}
\label{sec:section3.1}

We denote the class weight tensor of the final prediction layer ($w^{[p]}$) with ($[p]$) the layer index. We use Softmax to identify the class $c$ with the maximum probability and find corresponding class weight vector ($w^{[p]}_{c}$), denoted with red in Figure~\ref{fig:fig2}. To discover how each feature value of the selected weight vector can affect the final predictions, we perform a channel-wise multiplication between the prediction layer activation map ($a^{[p]}$) and the discovered class weight tensor. The produced class-based activation map ($a^{\star[p]}_{c}$) has the same dimensionality as the initial input activation map. However, each of the new features is solely based on the selected class (${[c]}$). A min-max feature scaling is applied to the activations in order to have a probabilistic distribution over the channel and feature activations, for each feature index ($i \in \{0,...,\mathit{d}\}$):

\begin{equation}
\label{eq:eq1}
\begin{split}
\bar{a}^{\star[p]}_{c} = \frac{a^{\star[p]}_{c,i} - min(a^{\star[p]}_{c})}{max(a^{\star[p]}_{c}) - min(a^{\star[p]}_{c})} \: \forall \:
i \in \{0,...,\mathit{d}\}
\end{split}
\end{equation}

In order to use globally aggregated class-based information, normalized over a probabilistic distribution ($\bar{a}^{\star[p]}_{c}$), we proceed with a third operation that explores the channel-wise dependencies of the class and the features extracted in a particular layer. This operation needs to be able to deal with both small probabilistic feature distributions, in the case of visually similar classes, and broader probability distributions, for easily distinguishable cases. Additionally, we need to be able to highlight multiple features in contrast to focusing on a single value in a one-hot fashion. These requirements led to the use of a simple monotonic shifted logistic sigmoid function, given a user-defined threshold value ($\theta$), such as in Equation~\ref{eq:eq2}. This also helps to mitigate the overall complexity of the method and increases the computational efficiency. The parameterization of the gate mechanism based on this threshold creates a bottleneck which can effectively reduce the pooled activation map's channel dimensionality to solely include high activations. We discuss the threshold parameter in Section~\ref{sec:section4.1}. 

\begin{equation}
\label{eq:eq2}
\begin{split}
feats_{i} = \{i \:: \: F^{[p]}_{i}>0 \}\: where \: \mathbf{F}^{[p]} = \frac{1}{1+\epsilon^{-x+\theta}}
\end{split}
\end{equation}

Through Equation~\ref{eq:eq2}, the indexes ($feats_{i}$) of the most dominant class features can be identified. The features detected by the sigmoid function have a direct correspondence to those in the previous layer's output, which in turn are concatenated given a set of kernels ($k^{[l]}$). Therefore, each dimension in the activations volume is directly related to a specific kernel. The influence of this specific kernel on the final class prediction is based on the aforementioned logistic function, shown in Figure~\ref{fig:fig2} with the orange cross-layer connections.

\subsection{Cross-layer feature dependencies}
\label{sec:section3.2}

As seen in Figure~\ref{fig:fig2}, the challenge when back-stepping through earlier time-space features extracted by the network, is the complexity of these features and their correspondence to class features used in predictions. This relates to the curse of dimensionality as there is no straightforward approach to represent higher-dimensional signals (such as those found in deeper layer activations) to a lower-dimensional space ({e.g.} early network layers) and vice versa. Even in the context of maintaining the same dimensional space in terms of size, as the operations performed are followed by non-linearities, the problem of cross-layer feature correspondence persists as the feature space representing information is different across pairs of non-linear layers. An additional problem specific to CNNs is the strict locality of their operations. Since a kernel's receptive field is determined by the layer number, with the input volume gradually decreasing in size through a forward pass, later layers tend to probe larger spatio-temporal patches with larger number of features. Therefore the feature region that each kernel is applied to, also varies in size.

In \textit{Class Feature Pyramids}, information is considered in a global manner, which implies that the localities of kernels are transformed in order to hold the accumulated information. This is done during the feature correspondence step which does not handle the detection of specific regions. Specifically, to back-step features from a convolution layer ($L$) to ($L-1$), based on the indexes ($feats_{i}$) in the activations, the corresponding kernels ($kernels^{[l]}_{j}$) are selected. Traversal through features in kernels and in the activation maps ($a^{[l]}$) is accomplished by pooling both vectors to sizes equal to that of their features:

\begin{equation}
\label{eq:eq3}
\begin{split}
a^{\star[l]} = \prod_{j=0}^{\mathit{d^{[l]}}}a'^{[l]} * w'^{[l]}_{j}, \: \forall \: j \in feats_{i} \\
where, \: a'^{[l]} = \mathit{Pool}(a^{[l]})\: and \:  w'^{[l]} = \mathit{Pool}(w^{[l]})
\end{split}
\end{equation}

As the created activation map ($a^{\star[l]}$) now correlates with the layer's class information in the absence of local features, the rest of the method is similar to how feature correspondence between layers is discovered for the class prediction layer in the beginning. The class-feature activation map ($a^{\star[l]}$) is normalized to a probabilistic distribution over the features (Equation~\ref{eq:eq1}). Based on the probabilistic class-feature vector, the features with the highest activations in layer ($L-2$) can be identified through the sigmoid activation function of Equation~\ref{eq:eq2}, in accordance with threshold value $\theta$.

\subsection{Layer-wise and feature-wise hierarchies}
\label{sec:section3.3}
An intrinsic part of the network's overall explainability is to examine the value of kernels of a specific layer ($L$) individually or as group of feature activations of a previous layer ($L-1$). The necessity for understanding the cross-layer activation dependencies becomes clear when visualizing shallow layers. Such layers have limited feature complexity and kernels is activated by many kernels in deeper layers. This one-to-many association impedes the creation of a coherent kernel dependency graph. To ensure a balanced view for connection across different layers, the activations found from the back-step process can be viewed both in individual kernels as separate concatenation of activations from the previous layer, as well as for the entire layer providing a concatenation of all informative kernels in the layer.

\textbf{Feature-wise kernel association:} Individual kernels in layer ($L$) can be represented as the average sum of activations from kernels in the previous layer ($L-1$). This is based on discovering the indexes of the most descriptive features in the scaled and pooled activation ($a^{\star[l]}_{k},\: where \: k \in kernels^{[l]}_{j}$) and performing a back-step to concatenate all activation maps in the previous layer that are part of these indexes. Then, the identified activation maps are averaged in order to be represented on top of the original input clip.

\textbf{Layer-wise feature relationships:} Supplementary to the singular kernel-oriented approach, layer-wise class activations aim towards a compact representation of the features associated with a specific class in an entire layer. In comparison to the kernel-based approach, and on top of their compact feature representation, layer-wise visualization also has the advantage that the number of occurrences of activations of specific kernels in the layer-wise volume is based on the number of connections that the filters have to features of previous layers. Therefore, the layer representation directly relates to the number of cross-layer connections, as informative features for multiple kernels in the following layer will have higher values than those with lower numbers of connections.

\subsection{Convolution block type invariance}
\label{sec:section3.4}

Visualizations produced by \textit{Class Feature Pyramids} are invariant to the convolution types and connections used. In comparison to other feature visualization methods \cite{szegedy2015going,szegedy2016rethinking,xie2017aggregated}, the nature of the convolution operation has no impact on our method and can be used even in architectures that include convolutions performed over the channel volumes, in parallel, and with varying kernel and channels dimensions performed at the same layer. In particular, our work addresses three different convolution operation types that exhibit a high degree of complexity in the way information is connected, and how they are performed. We illustrate our approach for these specific cases in Figure~\ref{fig:fig3}.

\textbf{Residual connections:} Spatio-temporal networks that employ residual connections have been widely used in video recognition \cite{diba2018spatio,hara2018can,qiu2019learning,tran2018closer}. In cases where bottleneck blocks are used, the back-step process becomes more complex as information is divided between two paths. This does not allow a straightforward hierarchical description of a backward pass through the network. We therefore include such cases in our method by creating tensors with values of one. The purpose of the created tensors is to act as a direct link between the proceeding and succeeding layers since the single value translates to an always-active state for the gatekeeper activation function. This translates to all the discovered feature indexes in the previous layer with high activations being passed through the next layer directly. Through this technique, activation indicxs are shared between previous network layers in the residual branch in the same manner as those in the main pathway of the block.

\textbf{Grouped convolutions:} Additional operations include the use of cross-channel convolutions performed in groups \cite{hara2018can,tran2019video}. These types of convolutions are especially challenging for back-stepping and visualization because their filters are of various channel depth. Consequently, there is no immediate association between the features that the filter uses and the total number of activations in the layer. This is dealt with by explicitly inflating each of the grouped kernels to correspond to the same dimensional space as the activation maps. This simulates how channel-wise operations are executed.

\textbf{Convolutions in branches:} The approach of decoupling information to multiple branches and streams has also been widely used for video classification \cite{carreira2017quo,chen2018multifiber,feichtenhofer2018slowfast,wang2018action}. Multiple pathways are created from the same activation maps, in which different operations occur. The variation in the type of operations and the number of operations adds a degree of ambiguity for constructing such blocks in a hierarchical manner. In these cases, back-steps through branches and pathways are accomplished with one-valued kernels and activation maps that act as small sub-structures that allow information to be passed directly across various paths.

\begin{figure}
\centering
\includegraphics[width=.9 \columnwidth]{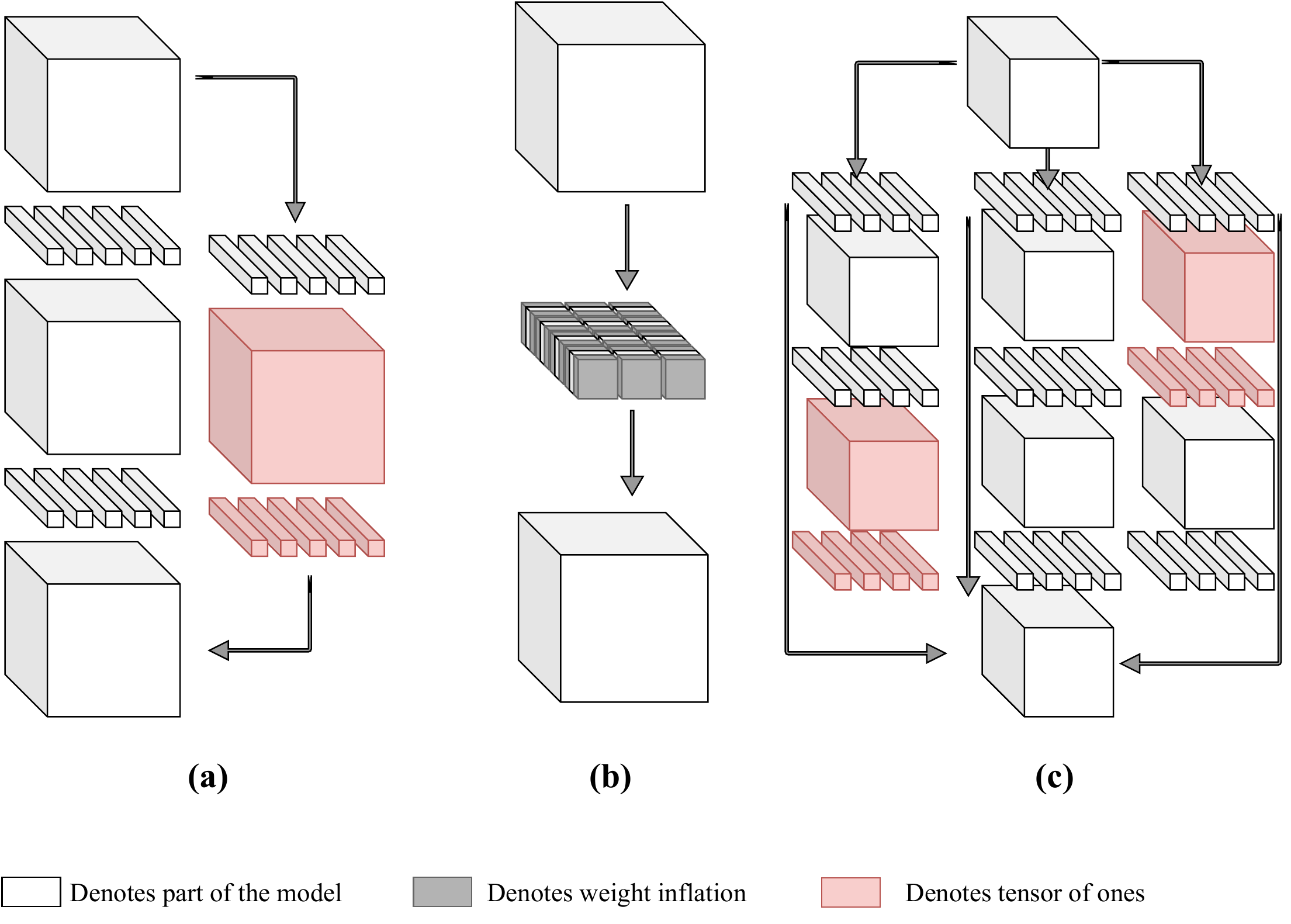}
\caption{\textbf{Back-step for different convolution types.} (a) \textbf{Residual connections}, kernels and activations of ones (\textcolor{red}{red}) are used to allow the discovery of channel associations with layers that are connected with residual connections. (b) \textbf{Grouped convolutions} in which the kernels are inflated (denoted with \textcolor{gray}{gray}) to hold the same dimensions as the input. (c) \textbf{Convolutions in branches} are performed in parallel with the possibility of uneven depth. In these cases, the branch with the maximum number of convolutions is selected as the base with tensors of ones added to the other branches.}
\label{fig:fig3}
\end{figure}

\section{Computational complexity and latency}
\label{sec:section4}

\textit{Class Feature Pyramids} is a generic visualization method that can be applied regardless of the overall complexity of the convolution operations or the between-layer connections. In this section, we describe the different spatio-temporal networks we employed to demonstrate our method and discuss the running times required for identifying the kernels with the highest activations per layer.

\subsection{Inference and running times}
\label{sec:section4.1}

We evaluate our approach on six different spatio-temporal networks to compare the times required during the back-step kernel search with different thresholds. The results are summarized in Table~\ref{table:table1}. We progressively increase the threshold value only to be proportional to the complexity of the model to ensure that a sufficient number of features is found for each architecture. Apart from the network architecture and the threshold value, latency also depends on the number of layers that class features are back-stepped to. This relates to features in earlier layers being reached slower than features in deeper layers. This is primarily based on the complexity of the considered features as more general features in early layers will have significantly more connections to high-layer features. Higher-layer features are more specific for certain classes.

\begin{table}[htb]
\begin{center}
\resizebox{\columnwidth}{!}{%
\begin{tabular}{l|c|c|c|c}
\hline
Network & GFLOPS & Back-step time (msec) & \# layers & $\theta$ \\
\hline
Multi-FiberNet \cite{chen2018multifiber} & 22.70 & 24.43 & 3 & 0.6 \\
\hline
I3D \cite{carreira2017quo} & 55.79 & 23.21 & 1 + mixed5c & 0.65 \\
\hline
ResNet50-3D \cite{hara2018can} & 80.32 & 21.39 & 3 & 0.55 \\
\hline
ResNet101-3D \cite{hara2018can} & 110.98 & 39.48 & 3 & 0.6 \\
\hline
ResNet152-3D \cite{hara2018can} & 148.91 & 31.06 & 3 & 0.6 \\
\hline
ResNeXt101-3D \cite{hara2018can} & 76.96 & 70.49 & 3 & 0.6 \\
\hline
\end{tabular}%
}
\end{center}
\caption[]{\textbf{Evaluation of running times}. The threshold value ($\theta$) is based on the model complexity. All architectures are back-stepped for three layers. For I3D, this corresponds to the class filters and the last \textit{mixed} block. All times are obtained on a machine with 2$\times$ Nvidia GTX 1080 Ti GPUs.}
\label{table:table1}
\end{table}

When creating class-based activation maps ($a^{\star[p]}_{c}$), the choice of global vectorized activations and features instead of performing in-layer convolutions is primarily attributed towards a significant reduction in operation time. The speed-up gained by using a global representation instead of the product of iterative local operations is proportional to the size of the spatio-temporal activation maps divided by the kernel's dimensions. Therefore, if we consider the input activations of size $D \times H \times W$ with $D$, $H$, and $W$ the number of frames, height and width, respectively, and the kernel sizes of $F_d \times F_h \times F_w$, the time required for iterating over the input activation map's channels changes from $O^{n \times (F_d/D \times F_h/H \times F_w/W)}$ to $O^n$ with vectorized volumes.

\subsection{Hierarchical visual interpretability outline}
\label{sec:section4.2}

We propose two ways to visualize features when generating network explanations. The first focuses on in-layer kernels being specific to the features extracted in individual layers. For the representation of each of the activations found, we create a spatio-temporal mask that contains the produced activation map of dimension ($i$). In the second approach, multiple class activations are stacked together to form a multi-feature visualization for the class layer features. To deal with the curse of dimensionality, in both cases, we reshape the volume through a polynomial spline interpolation and thus create a new activation map. More specifically, we define each representation of the activation volume as a multi-dimensional point. For each pair of produced points, we define a function $f_{(x,y)}$ where $(x,y)$ denotes the spatial extend of each frame in the clip and we find in-between points through a piece-wise polynomial function $S_{(x,y)}$ composed of an $n$-degree polynomial. For each activation, $n$ is equal to $T/t$, where $t$ is the temporal extend of the activation map and $T$ is the temporal extend of the original clip. The function $S_{(x,y)}$ is defined as:

\begin{equation}
\label{eq:eq4}
\begin{split}
\scalebox{0.8}{%
$S(x,y) = P_i(x,y) \; (x_{i-1},y_{i-1}) < (x,y) < (x_{i},y_{i})\; \forall 0 \leq i\leq n$
}
\end{split}
\end{equation}

\section{Visualization results}
\label{sec:section5}

In this section, we visualize the outputs of the back-step process as they are created from the representative layers of the network given the threshold from Equation~\ref{eq:eq2}, down to the selected layer or selected kernels of a layer. We explain how the proposed method can be used to visualize class features of different layers from various model architectures in two different video recognition tasks. Results on different types of class features extracted by six network architectures and 3D convolution operation variations are presented in Section~\ref{sec:sec5.1}. All models are evaluated on Kinetics-400, UCF-101 and HMDB-51. In Section~\ref{sec:sec5.2} we present instances of class features for egocentric actions for EPIC-Kitchens and EGTEA Gaze+ that demonstrate how kernel groups of the network distinctively follow parts of the scene.

\begin{figure*}[htb]
    \begin{minipage}[b]{0.15\textwidth}
        \bigbreak
        \bigbreak
        Multi-FiberNet\\
        \bigbreak
        \bigbreak
        I3D\\
        \bigbreak
        \bigbreak
        ResNet50-3D\\
        \bigbreak
        \bigbreak
        ResNet101-3D\\
        \bigbreak
        \bigbreak
        ResNet152-3D\\
        \bigbreak
        \bigbreak
        ResNeXt101-3D\\
        
    \end{minipage}
    \hfill
    \begin{minipage}[t]{.25\textwidth}
    \centering
        \animategraphics[loop,autoplay, controls, width=\textwidth]{8}{kinetics_3d_cnns/frame_}{0000}{0031}
        \captionsetup{labelformat=empty}
        \caption{\textbf{Kinetics-400}}
        \label{fig:fig4a}
    \end{minipage}
    \hfill
    \begin{minipage}[t]{.25\textwidth}
    \centering
        \animategraphics[loop,autoplay, controls, width=\textwidth]{8}{ucf101_3d_cnns/frame_}{0000}{0031}
        \captionsetup{labelformat=empty}
        \caption{\textbf{UCF-101}}
        \label{fig:fig4b}
    \end{minipage}
    \hfill
    \begin{minipage}[t]{.25\textwidth}
    \centering
        \animategraphics[loop,autoplay, controls, width=\textwidth]{8}{hmdb51_3d_cnns/frame_}{0000}{0031}
        \captionsetup{labelformat=empty}
        \caption{\textbf{HMDB-51}}
        \label{fig:fig4c}
    \end{minipage} 
    \setcounter{figure}{3}
\caption{\textbf{Kernel visualizations for action recognition datasets.} Each row of three frames corresponds to activations of different kernels at different depths. We define the layer depth of each kernel, its indexes and the number of connections that the kernels has to the preceding layer in the supplementary material. We use different threshold values for each of the networks. The first column corresponds to an example of \textit{building lego} in Kinetics-400. The threshold values used for the networks (row-wise) are [0.7, 0.7, 0.65, 0.65, 0.7, 0.7]. The second column is a \textit{rowing} clip from UCF-101 based on thresholds [0.8, 0.75, 0.7, 0.75, 0.8, 0.8]. Last example is from HMDB-51 of class \textit{ride bike} with the same thresholds as UCF-101.}
\label{fig:fig4}
\end{figure*}

\subsection{Class feature exploration in action recognition}
\label{sec:sec5.1}

In Figure~\ref{fig:fig4}, we show how features in different depths are visualized for different network architectures on the three action recognition datasets. We demonstrate the visual variations in class features observed considering how each network processes each clip. We examine the activations of spatio-temporal features across different depths of the network, rather than the regions with the highest activations for class weights. This means that the activations presented should include much more targeted regions and have significantly smaller duration. A full list of the kernels used can be found in the supplementary material.

In the first column of Figure~\ref{fig:fig4} the selected example is of class \textit{building lego} from the Kinetics-400 dataset. Feature activation visualizations between different networks vary significantly. In the case of the convolution operations that are grouped into fibers such as in MultiFiberNet, multiple small spatio-temporal regions are extracted at each layer because each of the kernels is performed in a sub-volume of the layer's activations. The same effect in a lesser extend is also present in both group-based and branch-based operations of the ResNeXt and I3D architectures, respectively. In contrast, when using models with convolutions performed over the entire activation map per single operation (\textit{i.e.} kernels and activation maps have equal dimensions) the regions of single-class informative features are more evident. In the \textit{building lego} class, ResNet architectures focus much more on either the hand region when picking or combining lego pieces (strong temporal cues) or the pile of pieces (strong appearance cue).

The second column presents a clip of \textit{rowing} from UCF-101. In this example, we visualize kernels from early layers where the effects of convolution type are more notable. A direct comparison between MultiFiberNet, with each operation corresponding to 16 fibers, and ResNext, with 32 kernel groups, shows that both networks focus on multiple regions. However, based on their key architectural difference, fibers consider a part of the input activation maps though feature slices. This makes the product of each operation discrete, while in contrast grouped convolutions in ResNext iterate channel-wise in a single layer of the proposed ResNeXt block. For I3D, we visualize a layer inside a \textit{mixed branch} that includes up to four different regions each of which corresponds to the different paths inside the branch. In the case of Residual networks, the feature activation regions remain distinct to specific movements and appearance features.

For the final clip, we select an instance of \textit{ride bike} from HMDB-51. In terms of network depth, we demonstrate a combination between features across multiple depths. In MultiFiberNet, the first visualizations are from the last convolution layer of the network while the proceeding two feature activations are from earlier network layers. Similarly, for the grouped convolutions in ResNeXt, the middle representation is drawn from earlier networks layers and is therefore focusing on a large number of spatio-temporal regions. In contrast, the feature activations in the two side visualizations are for the final convolution layers that specifically focus on certain parts of the movement. The feature activations chosen for the I3D, and ResNet variants all are based on the last three convolution layers.

\subsection{Egocentric spatio-temporal feature regions}
\label{sec:sec5.2}

In Figure \ref{fig:fig5}, we show the input frame sequences, overlaid with the activations of the class corresponding kernels that are above \(\theta\) for a layer. Results are for the MultiFiberNet network. The first four rows contain samples for EPIC-Kitchens and the last row is a clip from EGTEA Gaze+.

In the first row, we show how the network follows a `cut' action. In the last kernel visualization it focuses on the area where the knife meets the carrot. Eventually, it follows its movement along with the hands. Kernels in deeper layers seem to specialize on specific parts of the frames or the content, for example the cutting board, the movements of the knife and the shoulders. Similarly, for the `open' action in the second row, the network has higher class feature activations when the lid opens and the two box pieces are distinguished. This is observable by the strong activations in that area. Evidently, the other kernels are following the same pattern of specialization in different areas spatially and temporally. 

Rows 3 and 4 show the same segment from EPIC-Kitchens but for a different action class. In this case, the ground truth label is 'insert' however the network misclassifies it for 'put'. In the third row we visualize activation maps for action 'insert', while in the fourth for 'put'. We argue that not only there is a semantic relationship between the false predictions, but the almost overlapping activated regions (rows 3 and 4, first two columns) indicate the connection between the classes in deeper network layers as well. That leads to the conclusion that the network understands the two classes from similar features and only differentiates from kernels that have importance below our threshold.

Finally, we add a visualization from EGTEA Gaze+ to show how a modified perspective does not change the visualized features. We show the features for class `wash', which is the ground truth for the segment. In the first and second columns, the network tries to capture the hands, the sink and the water. In the third column, it concentrates further on the water, seemingly associating it with washing. Finally, in deeper layers it primarily captures salient objects such as the faucet and the sink's boundaries.


\begin{figure}
\centering
    \begin{animateinline}[poster=0, loop, controls]{2}
        \multiframe{16}{I= 0+1}{
            \begin{tabular}{@{\hspace{0pt}}c@{\hspace{0.02\linewidth}}c@{\hspace{0.02\linewidth}}c@{\hspace{0.02\linewidth}}c@{\hspace{0.02\linewidth}}c@{\hspace{0pt}}}
                \raisebox{0.065\linewidth}{cut}&
                \includegraphics[width=0.18\linewidth]{figure5/cut_long/l1_k5_185/frame_00\I.jpg} &
                \includegraphics[width=0.18\linewidth]{figure5/cut_long/l2_k24_5/frame_00\I.jpg} &
                \includegraphics[width=0.18\linewidth]{figure5/cut_long/l3_k547_24/frame_00\I.jpg} &
                \includegraphics[width=0.18\linewidth]{figure5/cut_long/l3_k557_40/frame_00\I.jpg} \\
                \raisebox{0.065\linewidth}{open}&
                \includegraphics[width=0.18\linewidth]{figure5/open/l1_k2_8/frame_00\I.jpg} &
                \includegraphics[width=0.18\linewidth]{figure5/open/l2_k73_2/frame_00\I.jpg} &
                \includegraphics[width=0.18\linewidth]{figure5/open/l3_k712_9/frame_00\I.jpg} &
                \includegraphics[width=0.18\linewidth]{figure5/open/l3_k719_4/frame_00\I.jpg} \\
                \raisebox{0.065\linewidth}{insert}&
                \includegraphics[width=0.18\linewidth]{figure5/insert/l1_k22_18/frame_00\I.jpg} &
                \includegraphics[width=0.18\linewidth]{figure5/insert/l2_k11_3/frame_00\I.jpg} &
                \includegraphics[width=0.18\linewidth]{figure5/insert/l3_k698_2/frame_00\I.jpg} &
                \includegraphics[width=0.18\linewidth]{figure5/insert/l3_k700_5/frame_00\I.jpg} \\
                \raisebox{0.065\linewidth}{insert-put}&
                \includegraphics[width=0.18\linewidth]{figure5/insert-put/l1_k1_2/frame_00\I.jpg} &
                \includegraphics[width=0.18\linewidth]{figure5/insert-put/l2_k214_3/frame_00\I.jpg} &
                \includegraphics[width=0.18\linewidth]{figure5/insert-put/l3_k457_19/frame_00\I.jpg} &
                \includegraphics[width=0.18\linewidth]{figure5/insert-put/l3_k459_354/frame_00\I.jpg} \\
                \raisebox{0.065\linewidth}{wash}&
                \includegraphics[width=0.18\linewidth]{figure5/wash-gtea/l1_k4_3/frame_00\I.jpg} &
                \includegraphics[width=0.18\linewidth]{figure5/wash-gtea/l2_k352_2/frame_00\I.jpg} &
                \includegraphics[width=0.18\linewidth]{figure5/wash-gtea/l2_k685_1/frame_00\I.jpg} &
                \includegraphics[width=0.18\linewidth]{figure5/wash-gtea/l3_k244_3/frame_00\I.jpg}
            \end{tabular}
        } 
    \end{animateinline}
\caption{\textbf{Kernel visualizations for egocentric datasets.} Visual results from MultiFiberNet on datasets EPIC-Kitchens (first four rows) and EGTEA Gaze+ (last row). From left to right, class activation maps from the last layer to gradually deeper ones. The threshold values for the visualizations are [0.84, 0.64, 0.84, 0.84, 0.84].}
\label{fig:fig5}
\end{figure}

\section{Conclusion}
\label{conclusions} \label{sec:section6}

Research on the explainability of spatio-temporal 3D-CNNs is limited, partly because of a scarcity of proper visualization methods. To overcome this deficit, we propose a lightweight, hierarchical approach named \textit{Class Feature Pyramids} that captures and presents informative features over different layers that are specific for a class. Our method is independent of the network type and can be employed regardless of the type of 3D convolution operation. Additionally, it enables the visualization of activations in layer-wise, group-wise or kernel-wise format. Our method is therefore suitable to visualize and, subsequently, to better understand what kind of features are learned to identify a specific class. This will aid in explaining the inner workings of 3D-CNNs, identifying potential biases of trained models, and to interpret the success or failure of a classification.

We have demonstrated the merits of our approach on six 3D-CNN architectures that include different connections and convolution operations. Our experiments are performed on five common third-person and egocentric action recognition datasets. The results provide insight into the learned features. For egocentric videos, they reveal attention towards both salient objects and the movement of the hands.

{\small
\bibliographystyle{ieee}
\bibliography{egbib}
}

\end{document}